\documentclass[numsec,webpdf,modern,large]{oup-authoring-template}
\usepackage{booktabs}
\usepackage{array}
\newcolumntype{L}[1]{>{\raggedright\arraybackslash}p{#1}}

\usepackage{pifont}

\usepackage{tabularx}
\usepackage{booktabs}

\graphicspath{{Fig/}}

\theoremstyle{thmstyleone}%
\theoremstyle{thmstyletwo}%
\theoremstyle{thmstylethree}%

\begin{document}

\journaltitle{Bioinformatics}
\DOI{DOI added during production}
\copyrightyear{YEAR}
\pubyear{YEAR}
\vol{XX}
\issue{x}
\access{Published: Date added during production}
\appnotes{Original Paper}

\firstpage{1}


\title[Short Article Title]{PVminerLLM2: Improving Structured Extraction of Patient Voice via Preference Optimization}

\author[1,2$\ast$]{Samah Jamal Fodeh\ORCID{0000-0003-4664-3143}}
\author[1]{Linhai Ma\ORCID{0000-0001-8519-864X}}
\author[1]{Ganesh Puthiaraju}
\author[1]{Srivani Talakokkul}
\author[1]{Afshan Khan}
\author[1]{Elyas Irankhah}
\author[1]{Sreeraj Ramachandran}
\author[3]{Ashley Hagaman}
\author[3]{Sarah Lowe}
\author[4] {Aimee Roundtree} 


\address[1]{\orgdiv{Department of Emergency Medicine}, \orgname{Yale School of Medicine}, \orgaddress{\street{464 Congress Ave}, \postcode{06519}, \state{CT}, \country{USA}}}

\address[2]{\orgdiv{Department of Biomedical Informatics \& Data Science}, \orgname{Yale School of Medicine}, \orgaddress{\street{100 College Street}, \postcode{06510}, \state{CT}, \country{USA}}}

\address[3]{\orgdiv{Department of Social and Behavioral Sciences}, \orgname{Yale School of Public Health}, \orgaddress{\street{60 College Street}, \postcode{06520}, \state{CT}, \country{USA}}}

\address[4]{\orgdiv{Division of Research}, \orgname{Texas State University}, \orgaddress{\street{601 University Dr.}, \postcode{78666}, \state{TX}, \country{USA}}}




\corresp[$\ast$]{Corresponding author. \href{samah.fodeh@yale.edu}{samah.fodeh@yale.edu}}

\received{Date}{0}{Year}
\revised{Date}{0}{Year}
\accepted{Date}{0}{Year}



\abstract{
\textbf{Motivation:}  
Patient-generated text contains critical information on patients’ lived experiences, social context, and care engagement, but remains largely unstructured, limiting its use in patient-centered outcomes research. Prior work introduced the PV-Miner benchmark and PVMinerLLM models for structured extraction. However, supervised fine-tuning (SFT) alone struggles with rare, fine-grained, and unevenly distributed errors, particularly in token-critical structured outputs. \\ 
\textbf{Results:}  
We present PVminerLLM2, an improved set of LLMs for structured patient voice extraction that applies preference optimization to address token-critical errors beyond the reach of supervised fine-tuning. Our method introduces (i) a preference objective with a token-level gated stabilization term that prevents degradation of absolute token likelihood under preference optimization, and (ii) confusion-aware preference pair construction to better capture low-separation distinctions. We further incorporate token-importance weighting and inverse-frequency reweighting to address token imbalance and class skew. Across multiple model sizes, PVMinerLLM2 consistently outperforms strong baselines, achieving gains of up to 4.43\% (Code), 3.50\% (Sub-code), and 1.55\% (Span), and outperforms baseline LLM trained with existing preference optimization methods.  \\
\textbf{Availability and Implementation:}  
The supplementary material, code, evaluation scripts, and trained models for PVminerLLM2 are publicly available at: https://github.com/Data-Mining-Lab-Yale/PVminerLLM2.  \\
\textbf{Contact:} \href{samah.fodeh@yale.edu}{samah.fodeh@yale.edu}  
}


\maketitle

\section{Introduction}

\begin{figure}[t]
\centering
\includegraphics[width=\columnwidth]{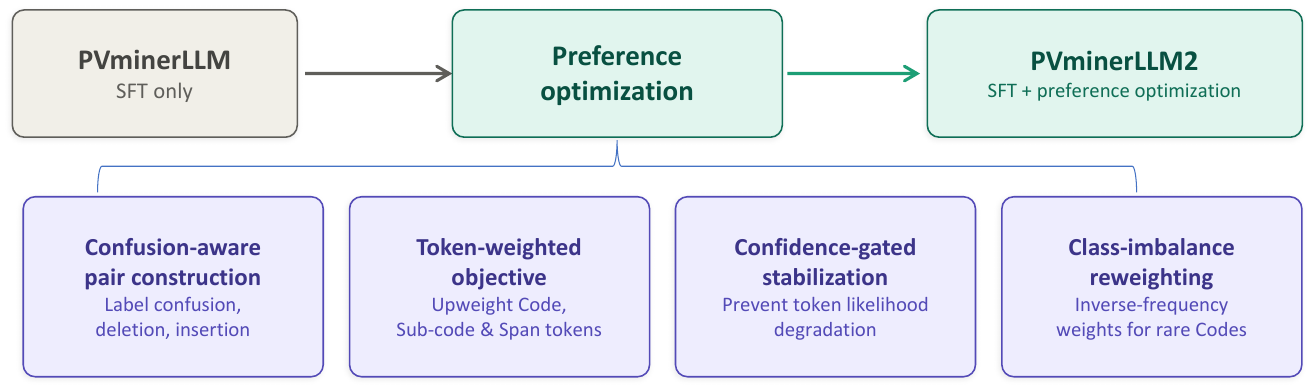}
\caption{Overview of the PVminerLLM2 training pipeline.
PVminerLLM (SFT only) is extended via preference optimization,
which incorporates four mechanisms: token-weighted objective,
confusion-aware pair construction, confidence-gated stabilization,
and class-imbalance reweighting.}
\label{fig:po_flowchart}
\end{figure}

Patient-generated data, including secure messages, survey responses, and interview narratives, provide a direct view into patients’ lived experiences outside traditional clinical encounters \citep{intro1, intro2, intro3, intro4}. Unlike structured clinical records, these texts capture how individuals describe their needs, constraints, emotions, and expectations in their own words. These expressions collectively form the \emph{patient voice}, reflecting not only clinical concerns but also broader social, environmental, and relational factors that influence health outcomes, such as social determinants of health and patient engagement in care \citep{intro5, intro6, intro7, fodeh2026pvminer, fodeh2026eppcminerben, fodeh2026tab, fodeh2026eppcoasis, fodeh2026star}.

Recent work \citep{fodeh2026pvminerllm} introduced the PV-Miner benchmark and showed that supervised fine-tuning (SFT) substantially improves over prompting-based approaches. However, SFT alone is insufficient: even small token-level errors, such as an incorrect Code or Sub-code label or a slight Span boundary deviation, can invalidate an otherwise well-formed output and cause downstream errors in clinical applications. These errors are rare, fine-grained, and non-uniform in severity, making them difficult to eliminate through likelihood-based training alone.

Preference optimization provides a complementary signal by explicitly comparing preferred and dispreferred structured outputs for the same input, enabling models to resolve confusable label assignments, reduce spurious or missing extractions, and improve Span grounding. However, standard sequence-level methods (e.g., \citep{rafailov2023direct,gheshlaghiAzar2024psiPO,meng2024simpo,xiao2024caldpo,pal2024smaug}) are insufficient here. In PV-Miner, preferred and dispreferred outputs often differ in only a few schema-defining tokens, so sequence-level objectives dilute the learning signal across largely identical outputs. Moreover, preference optimization can degrade the absolute likelihood of preferred outputs, harming structured prediction accuracy. Existing token-level methods \citep{zeng2024token,yang2025token} partially address credit assignment but do not stabilize absolute likelihood or target the construction of low-separation pairs.

To address these challenges, we present PVminerLLM2, an improved set of LLMs for structured
patient voice extraction that extends SFT with token-level preference optimization. Our proposed
method targets the three failure modes identified above: learning signal dilution, likelihood
degradation of preferred outputs, and class imbalance. We propose novel mechanisms at
both the objective and data-construction levels, including confidence-gated stabilization and confusion-aware pair construction. To our knowledge, PVminerLLM2 is the first to
integrate these two mechanisms for structured
information extraction. We evaluate PVminerLLM2 on the PV-Miner benchmark against PVminerLLM
and multiple DPO-based preference optimization methods across four model sizes.
 
\paragraph{Contributions:}
\begin{enumerate}
    \item We introduce a confidence-gated token-level stabilization mechanism that prevents
    degradation of absolute token likelihood during preference optimization. Our ablation
    identifies this as the single most impactful component.
 
    \item We propose a confusion-aware preference pair construction strategy that generates
    low-separation pairs from empirical SFT errors. This enables precise learning on
    token-critical distinctions.
 
    \item We incorporate token-importance weighting to concentrate learning on Code, Sub-code,
    and Span tokens. We further apply example-level inverse-frequency reweighting to address
    class imbalance.
 
    \item We provide a comprehensive empirical evaluation showing that PVminerLLM2 consistently
    outperforms PVminerLLM and existing DPO-based baselines across model sizes and evaluation
    levels.
\end{enumerate}

\section{Related Work}
\label{sec:related}

\begin{table*}[!h]
\centering
\caption{
Comparison of PVminerLLM2 preference optimization with existing methods. 
Prior approaches rely on completion-level signals, which can dilute learning under low edit-distance settings. 
PVminerLLM2 introduces token-level supervision and confidence-gated stabilization, focusing updates on semantically critical tokens (e.g., labels and spans).
}
\label{tab:po_comparison}
\small
\begin{tabularx}{\textwidth}{l c c c c c}
\toprule
\textbf{Method} 
& \textbf{Signal Level} 
& \textbf{Token Importance} 
& \textbf{Low Edit Dist.} 
& \textbf{Calibration / Stability} 
& \textbf{Protects Critical Tokens} \\
\midrule

DPO 
& Completion 
& No 
& No 
& Yes (ref reg.) 
& No \\

SimPO 
& Completion 
& No 
& No 
& Yes (implicit reward) 
& No \\

IPO / $\Psi$PO 
& Completion 
& No 
& No 
& Yes (margin regression) 
& No \\

Cal-DPO 
& Completion 
& No 
& No 
& Yes (calibration) 
& No \\

DPO-Positive 
& Completion 
& No 
& Partial 
& Yes (chosen likelihood) 
& No \\

\midrule

\textbf{
PVminerLLM2 (Ours)} 
& \textbf{Token + Completion} 
& \textbf{Yes (value-aware)} 
& \textbf{Yes} 
& \textbf{Yes (confidence-gated)} 
& \textbf{Yes (token-wise)} \\

\bottomrule
\end{tabularx}
\end{table*}

\subsection{DPO and its variants}
DPO \citep{rafailov2023direct} learns from preference pairs by encouraging a model to assign higher likelihood to preferred outputs than to dispreferred ones. Given an input $x$ and a pair $(y^+, y^-)$, DPO defines a preference score based on log-likelihood differences and optimizes:
\begin{equation}
    L_{\text{DPO}} = -\log \sigma\big(\beta (s^+ - s^-)\big),
\end{equation}
where $s^+ = \log \pi_\theta(y^+|x)$ and $s^- = \log \pi_\theta(y^-|x)$. Here, $\sigma(\cdot)$ denotes the sigmoid function, and $\beta$ is a scaling factor controlling the sharpness of the preference. In practice, DPO often incorporates a reference model (e.g., a supervised fine-tuned model) to stabilize training by comparing relative likelihoods rather than absolute values.

Building on DPO, several variants have been proposed. SImPO \citep{meng2024simpo} replaces the explicit reference model with an implicit reward; IPO/$\Psi$PO \citep{azar2024general} substitutes a margin-based regression objective; Cal-DPO \citep{xiao2024caldpo} adds a calibration mechanism; and DPO-Positive emphasizes the absolute likelihood of preferred outputs. However, all operate at the completion level and weight every token uniformly, leaving them without mechanisms to concentrate learning on the few schema-defining tokens that determine correctness in structured extraction. Our approach addresses this gap through token-level supervision and confidence-gated stabilization. See Table~\ref{tab:po_comparison} for a summary comparison and the \textbf{Extended DPO variant discussion} in supplementary materials for an extended discussion.

\subsection{Task and Data}
\label{task_definition}

\begin{table*}[t]
\centering
\caption{Example of patient voice extraction from the PV-Miner dataset. Each annotation is a (Code, Sub-code, Span) triple shown in JSON-style format.  ``None'' means the Code has no Sub-codes.}
\label{tab:examples}
\small
\setlength{\tabcolsep}{4pt}
\begin{tabular}{p{1.0cm} p{5.0cm} p{10.0cm}}
\toprule
Source & Message (Input) & Extracted Triples (Output) \\
\midrule
YNHH
  & Do I call them or do they reach out when they get the referral?
  & \texttt{[\{"code": "PartnershipPatient", "subcode": "activeParticipation",} \par
    \texttt{\ \ "span": "Do I call them"\},} \par
    \texttt{\ \{"code": "SharedDecisionPatient", "subcode": "ExploreOptions",} \par
    \texttt{\ \ "span": "or do they reach out"\},} \par
    \texttt{\ \{"code": "CareCoordinationPatient", "subcode": "None",} \par
    \texttt{\ \ "span": "they get the referral"\}]} \\
\bottomrule
\end{tabular}
\end{table*}

\begin{figure}[!h]
  \centering
  \begin{minipage}{0.45\columnwidth}
    \includegraphics[width=\linewidth]{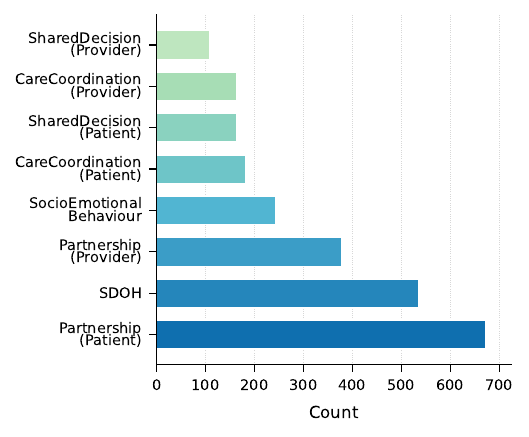}
  \end{minipage}
  \hfill
  \begin{minipage}{0.51\columnwidth}
    \includegraphics[width=\linewidth]{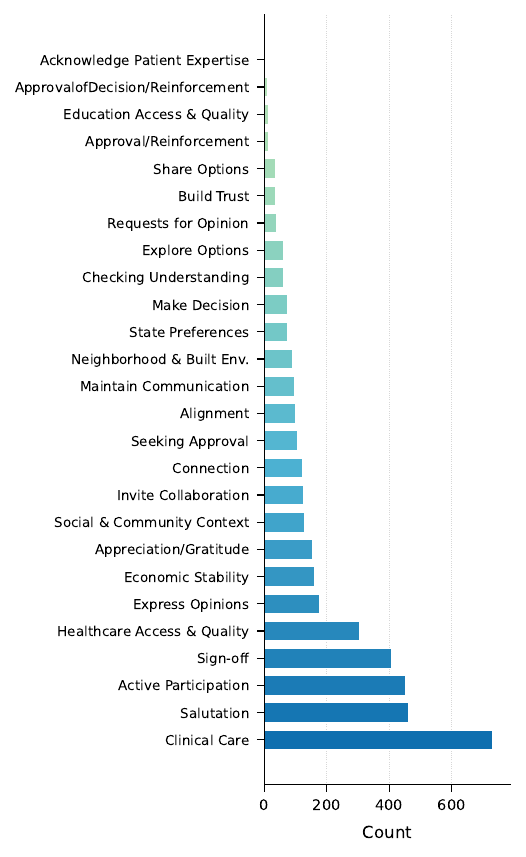}
  \end{minipage}
\caption{Distribution of Codes (left) and Sub-codes (right). The image is in high resolution, please zoom in for better vision.}
\label{fig:data_dist}
\end{figure}

We evaluate our method on patient generate datasets introduced in \citep{fodeh2026pvminerllm} called PV-Miner dataset, which consists of text data collected from multiple sources, including secure messages, surveys, and interview transcripts. The dataset contains 1,137 messages annotated with a hierarchical schema comprising 8 Codes and 33 Sub-codes, each paired with supporting text spans. In total, the dataset includes 46,038 words, with an average message length of 40.5 words (standard deviation 32.8), ranging from short clarifications to longer narrative descriptions. The data is inherently imbalanced, with a small number of high-frequency categories and a long tail of rare labels, reflecting real-world clinical communication patterns (shown in Fig. \ref{fig:data_dist} ). We follow the original train/test splits. 

Patient voice extraction is a structured information extraction task over patient-generated text. Given an input message, the goal is to extract all patient-centered communication elements as structured tuples of the form (Code, Sub-code, Span). Each tuple consists of a high-level category (Code), a more specific subtype (Sub-code), and the exact text span supporting the annotation. The task is multi-label, allowing multiple tuples per message, and is subject to schema constraints, where only valid Code--Sub-code combinations are permitted. Please find the \textbf{Codebook} in the supplementary materials.

Formally, given an input text $x$, the model predicts a set of tuples 
\begin{equation}
    \mathcal{Y}(x) = \{(c_i, s_i, t_i)\}_{i=1}^N, 
\end{equation}

where $c_i$ denotes a Code, $s_i$ is a Sub-code constrained by $c_i$, and $t_i$ is an exact span copied from the input text. We follow the task formulation, dataset construction, and annotation schema introduced in \citep{fodeh2026pvminerllm}, and refer readers to that work for full details. In practice, model outputs are generated in a structured JSON format to ensure schema validity and enable automatic evaluation. All datasets are in English. The messages were collected from patient portals and survey platforms where English is the primary communication language. Table \ref{tab:examples} presents one example. Examples from all the fours sources can be found in \textbf{Examples from all four data sources} in supplementary materials.

\section{Method}

\subsection{System overview}

We aim to improve preference optimization for structured extraction, where small token-level errors can invalidate the entire output. Our method extends standard DPO in three aspects: (i) focusing learning on task-critical tokens, (ii) stabilizing preferred outputs during training, and (iii) addressing class imbalance. In the training data perspective, we propose confusion-aware preference training data generation to better capture realistic errors.

\subsection{Token-weighted preference objective}
\label{sec:token_pref}

Following standard DPO, we optimize a preference objective based on likelihood differences between preferred and dispreferred outputs. However, in structured extraction, most tokens (e.g., formatting tokens in JSON outputs) do not affect correctness. Errors are typically localized to a small subset of tokens, such as Code, Sub-code, and Span values. Treating all tokens equally therefore dilutes the learning signal.

To address this, we redefine the sequence score as a token-weighted sum:
\begin{equation}
\label{eq:basic_loss}
s_\theta(y) = \sum_{t} w_t \log \pi_\theta(y_t \mid x, y_{<t}),
\end{equation}
where the weights depend on token roles:
\begin{equation}
w_t = w_{\text{Code}}\mathbb{I}[t \in \mathcal{C}] + w_{\text{Sub-code}}\mathbb{I}[t \in \mathcal{S}] + w_{\text{Span}}\mathbb{I}[t \in \mathcal{P}],
\label{eq:token_weights}
\end{equation}
where $\mathcal{C}$, $\mathcal{S}$, and $\mathcal{P}$ denote the sets of token positions that fall within Code, Sub-code, and Span values in the serialized output, respectively. Tokens outside all three sets (e.g., JSON scaffolding) receive $w_t = 0$; tokens belonging to multiple fields receive correspondingly higher weight.


Then, we compare the model against a reference model (the supervised fine-tuned (SFT) model). Let $s_{\text{ref}}(y)$ denote the same token-weighted score computed using this reference model.
\begin{equation}
\Delta =
\big(s_\theta(y^+) - s_\theta(y^-)\big)
-
\big(s_{\text{ref}}(y^+) - s_{\text{ref}}(y^-)\big),
\end{equation}
During the training process, the following objective is optimized:
\begin{equation}
L_{\text{pref}} = -\log \sigma(\beta \Delta).
\end{equation}

This reference-adjusted objective ensures that learning focuses on improvements over the SFT baseline rather than inheriting its biases.

\subsection{Confidence-gated likelihood stabilization}

Preference optimization may reduce the absolute likelihood of preferred outputs, which can degrade structured formats or span accuracy. To mitigate this, we introduce a confidence-gated stabilization mechanism.

We first identify unstable tokens using:
\begin{equation}
g_t =
\begin{cases}
1 & \text{if } \log \pi_\theta(y_t^+ \mid x, y_{<t}^+) < \gamma \\
0 & \text{otherwise}
\end{cases}
\end{equation}

which activates only for tokens whose predicted probability falls below a confidence level. Here, $\gamma$ controls the minimum confidence required for a token to be considered stable.

We then apply a weighted penalty:
\begin{equation}
L_{\text{stab}} =
\frac{
\sum_t g_t \, w_t \big(-\log \pi_\theta(y_t^+ \mid x, y_{<t}^+)\big)
}{
\sum_t g_t \, w_t
}.
\end{equation}

This term selectively lowers the preference optimization loss for low-confidence tokens in preferred outputs, preventing degradation while avoiding unnecessary constraints on already stable tokens.

\subsection{Class-imbalance reweighting}

The dataset exhibits a long-tail distribution (Fig.~\ref{fig:data_dist}), where rare Codes receive limited training signal. To address this, we adopt a class-balanced weighting strategy that increases the contribution of examples containing underrepresented categories.

Specifically, let $n_\kappa$ denote the frequency of Code $\kappa$ in the training data. For each training example $i$, we assign an inverse-frequency weighting based on the least frequent Code it contains:
\begin{equation}
\omega_i = \max_{\kappa \in \mathcal{C}_i} \frac{1}{n_\kappa},
\end{equation}
where $\mathcal{C}_i$ denotes the set of Codes in the example $i$.

This formulation ensures that examples involving rare Codes receive higher weights, while frequent ones are down-weighted. In practice, we use the effective-number formulation \citep{cui2019classbalanced} as a smooth variant of this inverse-frequency weighting.

The final training objective is defined as:
\begin{equation}
L =
\frac{\sum_i \omega_i \big(L_{\text{pref}}^{(i)} + \lambda L_{\text{stab}}^{(i)}\big)}{\sum_i \omega_i},
\end{equation}
where $L_{\text{pref}}$ is the preference loss and $L_{\text{stab}}$ is the stabilization loss.

\subsection{Preference Pair Construction}
\label{sec:pref_data}
Because many PV-Miner Codes and Sub-codes are semantically overlapping, effective preference
pairs must be low-separation: preferred and dispreferred outputs differ in only a small number
of schema-defining tokens. Confusion-aware synthetic pairs are derived from empirical SFT
errors on a held-out validation set. For each gold annotation $Y^+$, three dispreferred outputs
$Y^-$ are generated by applying three perturbations, each schema-validity-preserving:
(i) \textbf{label confusion}: a Code or Sub-code is replaced with a commonly confused
alternative while the Span is preserved; (ii) \textbf{missing annotation}: one tuple is deleted
to simulate under-extraction; (iii) \textbf{extra annotation}: a spurious tuple is inserted to
simulate over-extraction. Because $Y^+$ and $Y^-$ share the same JSON scaffold and differ in
only a few semantic tokens, these pairs directly instantiate the low-edit-distance regime
motivating the token-weighted objective in Section~\ref{sec:token_pref}. The \textbf{Preference
Pair Construction Details} can be found in supplementary materials.

\section{Experiments}

\subsection{Experimental Setting}

All benchmark experiments are conducted using lm-eval \citep{finben, eval-harness} with a vLLM backend for fast, reproducible generation.
We evaluate instruction-tuned LLMs spanning 1.5B--70B parameters, including Llama-3.3-70B-Instruct, Llama-3.1-8B-Instruct, Llama-3.2-3B-Instruct, and Qwen2.5-1.5B-Instruct.
For all models, we apply each model's native chat template and use deterministic decoding (temperature $=0$, no sampling) to produce schema-constrained JSON outputs. The full prompt is provided in \textbf{Prompt for Training the LLM} in the supplementary materials.


\subsubsection{Metric}\label{append:metric}

Code and Sub-code prediction is evaluated as multi-label classification. Let $\hat{y}_i$ and $y_i$ denote the predicted and gold code sets for instance $i$. Micro-averaged precision, recall, and F1 are:

\begin{equation}
    precision=\frac{\sum_{i}\left|\hat{y}_{i} \cap y_{i}\right|}{\sum_{i}|\hat{y}_{i}|}
\end{equation}
\begin{equation}
    recall = \frac{\sum_i \left| \hat{y}_i \cap y_i \right|}{\sum_i \left| y_i \right|}
\end{equation}
\begin{equation}
    F1 = \frac{2 \times precision \times recall}{precision + recall}
\end{equation}

For Span extraction, a predicted Span is a true positive (TP) if it fully contains, or is fully contained by, a gold Span, or if their token-level Jaccard similarity exceeds 0.6. Unmatched predictions are false positives (FP); unmatched gold Spans are false negatives (FN). Span F1 is computed analogously to Code F1 above.






\subsubsection{Hyperparameters}

Evaluation uses zero-shot decoding (temperature $=0$, max 8096 context tokens, max 1024 generated tokens) with schema-valid JSON output. Training uses QLoRA \citep{dettmers2023qlora} with bfloat16 mixed precision, 4-bit quantization, gradient checkpointing, and AdamW via HuggingFace \texttt{Trainer} on two H200 GPUs. Preference optimization is initialized from the publicly released PVminerLLM checkpoints (Fodeh et al., 2026a), which also serve as the frozen reference policy in Eq. \ref{eq:basic_loss}. SFT training details are described in the original work \citep{fodeh2026pvminerllm}. For token weighting, we upweight semantic fields within the JSON completion with $(w_{\text{Code}},w_{\text{Sub-code}},w_{\text{Span}})=(2,3,1.5)$, additionally upweighting tokens that differ between chosen/rejected by a factor of 2, and normalizing weight mass so that the mean active weight is approximately 1.0. $\gamma$ is 0.66 and $\beta$ is 0.85. 3 epochs of training is conducted. 

\subsection{PVminerLLM vs PVminerLLM2}

Table~\ref{tab:merged_results_std_improve} reports extraction performance across four model sizes at Code, Sub-code, and Span levels. PVminerLLM2 outperforms PVminerLLM on all models and all evaluation levels, with gains ranging from $+0.59\%$ to $+4.43\%$ F1 on Code, $+1.42\%$ to $+3.50\%$ on Sub-code, and $+0.74\%$ to $+1.55\%$ on Span. The largest absolute gains occur on Qwen2.5-1.5B ($+4.43\%$ Code, $+3.50\%$ Sub-code), while the smallest occur on Llama-3.3-70B ($+0.59\%$ Code). Alongside F1 improvements, PVminerLLM2 consistently reduces run-to-run variance: Qwen2.5-1.5B's Code F1 standard deviation drops from 2.71 to 0.26, and Llama-3.3-70B's Sub-code standard deviation from 2.40 to 0.35.

\begin{table}[t]
\centering
\caption{Extraction performance (in \%) (F1, Precision, Recall, and F1 std). $\Delta$F1 denotes improvement from SFT to PO. F1 is the mean of 3 runs with different random seeds}
\label{tab:merged_results_std_improve}
\small
\setlength{\tabcolsep}{3pt}
\begin{tabular}{lcccccccccc}
\toprule
& \multicolumn{4}{c}{PVminerLLM} & \multicolumn{4}{c}{PVminerLLM2} & $\Delta$F1 \\
\cmidrule(lr){2-5} \cmidrule(lr){6-9}
Model & P & R & F1 & Std & P & R & F1 & Std &  \\
\midrule

\multicolumn{10}{l}{\textbf{Code}} \\
\midrule
Llama-3.3-70B & 87.90 & 80.11 & 83.82 & 0.62 & 87.26 & 81.74 & \textbf{84.41} & 0.17 & +0.59 \\
Llama-3.1-8B  & 85.04 & 78.12 & 81.43 & 0.60 & 84.46 & 81.56 & \textbf{82.98} & 0.36 & +1.55 \\
Llama-3.2-3B  & 82.48 & 78.30 & 80.33 & 0.78 & 82.70 & 80.66 & \textbf{81.67} & 0.19 & +1.34 \\
Qwen2.5-1.5B  & 83.19 & 71.61 & 76.97 & 2.71 & 85.63 & 77.58 & \textbf{81.40} & 0.26 & +4.43 \\

\midrule
\multicolumn{10}{l}{\textbf{Sub-code}} \\
\midrule
Llama-3.3-70B & 83.74 & 77.95 & 80.74 & 2.40 & 84.13 & 80.34 & \textbf{82.19} & 0.35 & +1.45 \\
Llama-3.1-8B  & 79.19 & 76.33 & 77.73 & 0.53 & 79.16 & 79.14 & \textbf{79.15} & 0.38 & +1.42 \\
Llama-3.2-3B  & 75.80 & 73.74 & 74.75 & 1.75 & 77.54 & 75.03 & \textbf{76.27} & 0.30 & +1.52 \\
Qwen2.5-1.5B  & 77.86 & 66.88 & 71.96 & 1.75 & 76.15 & 74.77 & \textbf{75.46} & 0.46 & +3.50 \\

\midrule
\multicolumn{10}{l}{\textbf{Span}} \\
\midrule
Llama-3.3-70B & 88.02 & 86.07 & 87.03 & 1.30 & 88.94 & 88.22 & \textbf{88.58} & 0.37 & +1.55 \\
Llama-3.1-8B  & 87.29 & 86.37 & 86.83 & 2.74 & 86.16 & 89.40 & \textbf{87.75} & 0.59 & +0.92 \\
Llama-3.2-3B  & 85.29 & 84.34 & 84.81 & 0.61 & 85.73 & 86.25 & \textbf{85.99} & 0.45 & +1.18 \\
Qwen2.5-1.5B  & 83.98 & 85.64 & 84.80 & 0.57 & 84.66 & 86.44 & \textbf{85.54} & 0.49 & +0.74 \\

\bottomrule
\end{tabular}
\end{table}

\subsection{Comparison with Other DPO Variants}
 
We compare PVminerLLM trained with DPO, IPO, Cal-DPO, and DPO-Positive against PVminerLLM2 on Qwen2.5-1.5B-Instruct (Table~\ref{tab:compare}). SimPO failed to converge under our training configuration and is therefore excluded. PVminerLLM2 achieves the highest F1 at all three evaluation levels. IPO yields the lowest performance (Code F1 54.48\%), falling well below the SFT baseline. DPO (70.24\%) and Cal-DPO (66.80\%) also underperform SFT, indicating that reference regularization and calibration alone are insufficient in this low-edit-distance setting. DPO-Positive (72.43\%) comes closest to PVminerLLM2 (81.40\%) among the baselines but still falls short by 8.97\% on Code, 6.39\% on Sub-code, and 0.63\% on Span. PVminerLLM2 also exhibits the lowest run-to-run variance on Code (0.26) and Sub-code (0.46).





\begin{table}[!h]
\centering
\caption{Performance comparison (F1 score, \%) across methods on Qwen2.5-1.5B-Instruct. Std is over 3 runs with different random seeds.}
\label{tab:compare}
\small
\setlength{\tabcolsep}{3pt}
\begin{tabular}{lcccccc}
\toprule
 & \multicolumn{2}{c}{Code} & \multicolumn{2}{c}{Sub-code} & \multicolumn{2}{c}{Span} \\
\cmidrule(lr){2-3} \cmidrule(lr){4-5} \cmidrule(lr){6-7}
Method & F1 & Std & F1 & Std & F1 & Std \\
\midrule
PVminerLLM (SFT)      & 76.97 & 2.71 & 71.96 & 1.75 & 84.80 & 0.57 \\
\midrule
PVminerLLM + IPO       & 54.48 & 1.74 & 48.27 & 5.17 & 68.01 & 1.65 \\
PVminerLLM + DPO       & 70.24 & 0.63 & 64.44 & 0.54 & 83.27 & 0.11 \\
PVminerLLM + Cal-DPO   & 66.80 & 0.29 & 65.76 & 1.60 & 82.90 & 2.15 \\
PVminerLLM + DPO-P     & 72.43 & 0.52 & 69.07 & 0.38 & 84.91 & 0.30 \\
\midrule
PVminerLLM2 (Ours)     & \textbf{81.40} & 0.26 & \textbf{75.46} & 0.46 & \textbf{85.54} & 0.49 \\
\bottomrule
\end{tabular}
\end{table}

\subsection{Ablation Study}

\begin{table}[!h]
\centering
\caption{PVminerLLM2 ablation study on Qwen2.5-1.5B-Instruct. Results are
F1 (\%). Mean is the average of Code/Sub-code/Span (in \%). CB: class
balance weighting; TW: token weighting; CG: confidence-gated. T: true.
F: false.}
\label{tab:tabpo_ablation}
\small
\setlength{\tabcolsep}{6pt}
\begin{tabular}{ccc|cccc}
\toprule
\textbf{CB} & \textbf{TW} & \textbf{CG} & \textbf{Code} & \textbf{Sub-code} & \textbf{Span} & \textbf{Mean} \\
\midrule
F & F & F & 70.24 & 64.44 & 83.27 & 72.65 \\
F & F & T & \textbf{77.88} & \textbf{74.35} & \textbf{84.13} & \textbf{78.79} \\
T & F & F & 72.43 & 67.52 & 84.10 & 74.68 \\
F & T & F & 74.97 & 68.01 & 83.07 & 75.35 \\
\midrule
T & T & T & 81.40 & 75.46 & 85.54 & 80.80 \\
\bottomrule
\end{tabular}
\end{table}

Table~\ref{tab:tabpo_ablation} reports the contribution of each component
on Qwen2.5-1.5B-Instruct. The base configuration (no CB, no TW, no CG)
achieves a mean F1 of $72.79\%$. Enabling CG alone yields the largest
gain ($+6.00\%$, to $78.79\%$), and achieves the highest score in every
column, confirming that confidence-gated stabilization is the most
critical mechanism. CB and TW contribute more modest improvements of
$+1.89\%$ and $+2.56\%$ respectively, reflecting the benefit of
 token-importance concentration on the
token-critical PV-Miner task. Because of the multi-label nature of the task, class-balance reweighting yields more modest gains for this long-tailed situation. (F, F, F) configuration is DPO with confusion-aware pairs but no TW, no CG, no CB and it falls below SFT, which further shows that the components are necessary.

\subsection{Class-level analysis on Qwen2.5-1.5B}
\label{sec:class_analysis}

\begin{table}[t]
\centering
\caption{Code-level performance comparison between PVminerLLM and PVminerLLM2 F1. $\Delta$F1 denotes improvement.}
\label{tab:code_level_detail}
\small
\setlength{\tabcolsep}{4pt}
\begin{tabular}{lcccc}
\toprule
Class & Freq & PVminerLLM & PVminerLLM2 & $\Delta$F1 \\
\midrule
SharedDecisionPatient & 163 & 31.11 & \textbf{59.02} & +27.91 \\
SocioEmotionalBehaviour & 242 & 63.46 & \textbf{79.66} & +16.20 \\
CareCoordinationPatient & 182 & 62.50 & \textbf{72.22} & +9.72 \\
SharedDecisionProvider & 108 & 42.42 & \textbf{45.83} & +3.41 \\
PartnershipProvider & 378 & 87.72 & \textbf{90.61} & +2.89 \\
PartnershipPatient & 671 & 85.71 & \textbf{88.41} & +2.70 \\
SDOH & 536 & 83.05 & \textbf{85.34} & +2.29 \\
CareCoordinationProvider & 162 & \textbf{77.50} & 72.73 & -4.77 \\
\bottomrule
\end{tabular}
\end{table}

\begin{table}[t]
\centering
\caption{Sub-code level performance comparison between PVminerLLM and PVminerLLM2 F1. $\Delta$F1 denotes improvement. Only Sub-codes with $\Delta$ F1 $>$ 10\% and $\Delta$F1 $<$ 0 \% are shown because the full list is long. The full table can be found in \textbf{Full Sub-code performance table} in supplementary materials.}
\label{tab:subcode_level_detail}
\small
\setlength{\tabcolsep}{4pt}
\renewcommand{\arraystretch}{0.95}
\begin{tabular}{P{2.2cm}cccc}
\toprule
Class & Freq & PVminerLLM & PVminerLLM2 & $\Delta$F1 \\
\midrule
checkingUnder\allowbreak standing/clarification
& 61 & 11.11 & \textbf{38.71} & +27.60 \\
Neighborhood\allowbreak AndBuiltEnvironment
& 91 & 28.57 & \textbf{48.65} & +20.08 \\
EducationAccess\allowbreak AndQuality
& 13 & 66.67 & \textbf{85.71} & +19.04 \\
requestsForOpinion
& 39 & 28.57 & \textbf{47.06} & +18.49 \\
ExploreOptions
& 61 & 14.29 & \textbf{31.58} & +17.29 \\
SeekingApproval
& 107 & 33.33 & \textbf{50.00} & +16.67 \\
connection
& 121 & 65.12 & \textbf{77.55} & +12.43 \\
SocialAnd\allowbreak CommunityContext
& 129 & 70.59 & \textbf{81.36} & +10.77 \\
expressOpinions
& 177 & 58.67 & \textbf{69.14} & +10.47 \\
signoff
& 406 & \textbf{92.74} & 91.02 & -1.72 \\
statePreferences
& 74 & \textbf{60.87} & 56.00 & -4.87 \\
EconomicStability
& 160 & \textbf{83.33} & 77.33 & -6.00 \\
build trust
& 37 & \textbf{50.00} & 33.33 & -16.67 \\
\bottomrule
\end{tabular}
\end{table}

Table~\ref{tab:code_level_detail} shows Code-level results on Qwen2.5-1.5B. The largest gains are on SharedDecisionPatient ($+27.91\%$, from $31.11\%$ to $59.02\%$) and SocioEmotionalBehaviour ($+16.20\%$, from $63.46\%$ to $79.66\%$). CareCoordinationPatient improves by $+9.72\%$; remaining Codes show smaller but consistent gains. The one regression is CareCoordinationProvider ($-4.77\%$).

Table~\ref{tab:subcode_level_detail} shows Sub-code results. The largest gains occur on low-frequency Sub-codes: checkingUnderstanding/clarification ($+27.60\%$), NeighbourhoodAndBuiltEnvironment ($+20.08\%$), EducationAccessAndQuality ($+19.04\%$), and requestsForOpinion ($+18.49\%$). Three Sub-codes regress: \textit{build trust} ($-16.67\%$), \textit{EconomicStability} ($-6.00\%$), and \textit{statePreferences} ($-4.87\%$).

\section{Discussion}

\subsection{Interpreting the Results}
\label{sec:discussion_results}

Sequence-level preference optimization methods spread the learning signal across hundreds of identical scaffolding tokens, providing almost no gradient on the schema-defining tokens that determine correctness. This explains why IPO collapses below SFT, and why DPO and Cal-DPO only partially recover through reference regularization and calibration: neither is token-selective. PVminerLLM2 succeeds by concentrating updates on Code, Sub-code, and Span tokens while the confidence-gated mechanism prevents their absolute likelihood from degrading under contrastive training. The ablation confirms that likelihood degradation, not insufficient learning signal, is the primary failure mode: without stabilization, even confusion-aware pairs with token weighting fall below SFT. Gains are larger on smaller models because they produce more frequent and diverse token-level errors after SFT, providing richer signal for confusion-aware pairs to exploit; larger models leave less room for correction. At the class level, the largest improvements occur on mid-to-low frequency Codes with high label confusion under SFT. The regressions on CareCoordinationProvider and build\_trust reflect a conservatism trade-off: confusion-aware pairs sharpen inter-label distinctions but can over-correct on closely related categories, as evidenced by build\_trust regressing while the semantically adjacent connection gains.

\subsection{Clinical and Social Implications}

PVminerLLM2 extends PVminerLLM with more reliable extraction of precisely the categories most relevant to health equity. The improved performance on SharedDecisionPatient, SocioEmotionalBehaviour, and rare SDOH Sub-codes means that signals of decisional uncertainty, emotional distress, and social barriers can now be detected at scale with substantially higher reliability. Recovering these signals enables systematic population-level monitoring of patient voice, supporting earlier identification of at-risk groups and more targeted care coordination. That the largest gains occur on smaller models further suggests that reliable patient voice extraction is accessible to resource-constrained health systems where these signals are most needed.

\subsection{Limitations and Future Work}

PVminerLLM2 demonstrates strong extraction capability but leaves room for improvement. The over-correction pattern identified in Section~\ref{sec:discussion_results} shows that confusion-aware pair construction can degrade performance on high-overlap classes; future work will investigate asymmetric margin strategies to mitigate this. The current prompt design is also necessarily complex due to the hierarchical schema and disambiguation requirements, and future work will explore multi-agent inference frameworks where distinct agents handle label selection and Span verification separately, which may reduce prompt complexity while improving robustness \citep{multi1, multi3, multi4}.

\section{Conclusion}

We present PVminerLLM2, an improved set of LLMs for structured patient voice extraction that applies preference optimization to address the token-critical, low-separation failure modes that SFT alone cannot resolve. PVminerLLM2 incorporates confusion-aware preference construction with confidence-gated token-level stabilization, and consistently improves over PVminerLLM and existing DPO-based baselines across model sizes, with especially strong gains on rare Sub-codes and reduced run-to-run variance. Together, the PV-Miner benchmark and PVminerLLM2 provide a practical testbed for studying preference optimization in real-world structured clinical NLP.

\section{Conflicts of interest}
The authors declare that they have no competing interests.



\section{Author contributions statement}

S.F. conceptualized the study, designed the methodology and data analysis plan, and contributed to the manuscript revision. L.M. designed the methodology, conducted the experiments, and prepared the manuscript. G.P. , A.K. , E.I. , S.R. assisted with the experiments, data processing, manuscript revision. S.T. and L.M. performed data annotation and quality assurance. S.F. , A.H. , S.L. , A.R. and S.T. provided domain expertise in patient-provider communication and contributed to the development and revision of the annotation framework. All authors reviewed and approved the final manuscript.

\section{Acknowledgments}
This work was supported by the Patient-Centered Outcomes Research Institute (PCORI) under Award No. ME-2023C2-31367 (to S.F.).

\bibliographystyle{oup-abbrvnat}
\bibliography{reference}

\end{document}